\pdfoutput=1

\documentclass[11pt]{article}

\usepackage{EMNLP2023}

\usepackage{times}
\usepackage{latexsym}

\usepackage[T1]{fontenc}

\usepackage[utf8]{inputenc}

\usepackage{microtype}

\usepackage{inconsolata}

\usepackage{graphicx}
\usepackage{hyperref}
\usepackage{url}
\usepackage{wrapfig}
\usepackage{booktabs}
\usepackage{multirow}
\usepackage{amsmath, amssymb}
\usepackage{subcaption}

\title{Learning to translate by learning to communicate}

\author{
  C.M. Downey$^{\alpha *}$ \quad
  Xuhui Zhou $^{\beta}$\thanks{~~Equal contribution. We also include a detailed Author Contribution Statement at the end of the paper.} \quad 
  Leo Z. Liu$^{\gamma}$ \quad
  Shane Steinert-Threlkeld$^{\alpha}$\\
  $^{\alpha}$Department of Linguistics, University of Washington \\
  $^{\beta}$Language Technologies Institute, Carnegie Mellon University \\
  $^{\gamma}$Department of Computer Science, The University of Texas at Austin\\
  {\tt \{cmdowney,shanest\}@uw.edu} \\
  {\tt zliu@cs.utexas.edu, xuhuiz@cs.cmu.edu}
}

\begin{document}

\maketitle

\begin{abstract}
  We formulate and test a technique to use Emergent Communication (EC) with a pre-trained multilingual model to improve on modern Unsupervised NMT systems, especially for low-resource languages. It has been argued that the current dominant paradigm in NLP of pre-training on text-only corpora will not yield robust natural language understanding systems, and the need for grounded, goal-oriented, and interactive language learning has been highlighted. In our approach, we embed a multilingual model (mBART, \citealp{liu-etal-2020-multilingual-denoising}) into an EC image-reference game, in which the model is incentivized to use multilingual generations to accomplish a vision-grounded task. The hypothesis is that this will align multiple languages to a shared task space. We present two variants of EC Fine-Tuning \citep{shane_etal_2022}, one of which outperforms a backtranslation-only baseline in all four languages investigated, including the low-resource language Nepali.
\end{abstract}

\section{Introduction}
\begin{figure}[t]
\centering
\includegraphics[width=0.95\columnwidth,trim=.4em 0.3em .4em .7em]{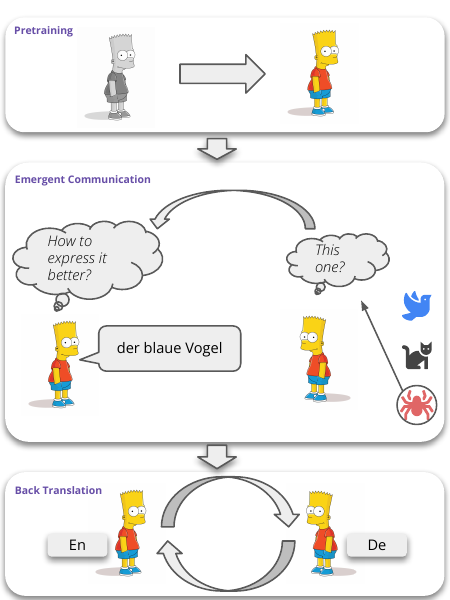}
\caption{Illustration of our modeling process. For the \textit{pre-training} stage, we use the off-the-shelf mBART \citep{lewis_bart_2020}. We fine-tune the model for translation with Emergent Communication.}
\label{fig:introFig}
\end{figure}
While neural machine translation (NMT) systems are one of the great success stories of natural language processing \cite{sutskever_sequence_2014, bahdanau_neural_2015, wu_googles_2016}, typical methods rely on large quantities of \emph{parallel text} (i.e.\ existing human translated texts) as gold data for supervised learning. These approaches are thus difficult to apply to low-resource languages, which lack large bodies of such data \cite{joshi_state_2020}. To extend this vital language technology to low-resource languages, many have focused on \emph{Unsupervised NMT} (UNMT) --- the task of building NMT systems without  any parallel text \cite{artetxe_unsupervised_2018, lample_unsupervised_2018, lample-etal-2018-phrase, lample_cross-lingual_2019, conneau-etal-2020-unsupervised}.

Typical approaches to UNMT rely on large pre-trained multilingual models \cite{lample_cross-lingual_2019, conneau-etal-2020-unsupervised, liu-etal-2020-multilingual-denoising, song_mass_2019} and the method of \emph{back-translation} \cite{sennrich-etal-2016-neural} to iteratively generate synthetic parallel text. These approaches, however, still rely on plain text information alone.  For that reason, the resulting models are considered \emph{ungrounded} (there is no link between the text and the external world). This may limit model abilities.

Despite NLP breakthroughs stemming from large-scale pre-training on raw text corpora with self-supervised learning \citep[ i.a.]{howard-ruder-2018-universal, peters-etal-2018-deep, devlin-etal-2019-bert, liu_roberta_2019, conneau-etal-2020-unsupervised, liu-etal-2020-multilingual-denoising, brown_language_2020}, several recent results suggest limitations in model generalization \citep[ i.a.]{mccoy-etal-2019-right, niven-kao-2019-probing, ettinger-2020-bert, rogers-etal-2020-primer}. More fundamentally, several have argued that pre-training on text alone will not deliver fully general and robust NLP systems.\footnote{This is largely what \cite{linzen-2020-accelerate} calls the pre-training Agnostic Independently Distributed (PAID) evaluation paradigm. We discuss pre-training on multimodal (i.e.\ not text-only) data in \S~\ref{sec:multimodal}.}

For example, using several detailed thought experiments, \citet{bender-koller-2020-climbing} argue that models trained on text alone will not, in principle, be able to recover either the conventional meaning of expressions or the communicative intent of an expression in context.  Their arguments highlight the importance of the interaction between linguistic expressions and extra-linguistic communicative intents (e.g.\ acting in the world, executing programs).\footnote{See \citet{merrill_provable_2021} for a formalization of argument in \citet{bender-koller-2020-climbing} about learning a programming language from form alone.} Similarly, \citet{bisk-etal-2020-experience} articulate progressively broader \textit{world scopes} in which language use is embedded, and argue that present pre-training methods work at a relatively limited scope.  They too emphasize the importance of embodied interaction with the environment and with the social world for future NLP systems.\footnote{As noted by \citet{bender-koller-2020-climbing}, many of these arguments can be seen as detailed elaborations of the need for NLU systems to solve the \emph{symbol grounding} problem \cite{harnad_symbol_1990, taddeo_solving_2005}.} 

In this paper, we propose to use methods from the field of \emph{emergent communication} (EC) \citep{wagner_progress_2003, skyrms_signals_2010, lazaridou_emergent_2020} to improve UNMT systems. EC studies artificial agents communicating with each other to accomplish particular environmental goals. EC is a subfield of reinforcement learning, wherein language (i.e.\ the communcation protocol) is shaped by rewards determined by interacting with an external environment and with other agents. Typical work in this area starts from a \emph{tabula rasa} and studies under what conditions ---e.g.\ environments, tasks/goals, social settings---the resulting communication protocols among agents resembles human language, along axes like word length economy \citep{chaabouni_anti-efficient_2019}, word-order biases \citep{chaabouni_word-order_2019}, and compositionality \citep{jacob_measure_2019, chaabouni_compositionality_2020,steinert-threlkeld_toward_2020,geffen_lan_spontaneous_2020}, among others \citep{mu_emergent_2021}.

Our approach leverages the insight that people learn new languages by using them to do things (e.g.\ order food, buy train tickets); our machines should do the same. We improve upon a standard UNMT system by taking a large pre-trained multilingual model (mBART) and embedding it in an EC task, having it participate in goal-directed communication (in addition to back-translation). Communication should promote translation in the following way. Translation can be viewed as `aligning' model representations for sentences in several languages. In the supervised case, parallel text instructs the model how to do this alignment. In the unsupervised case, through communication, each model aligns its language representations \emph{with the same shared environment}, thereby promoting alignment between the languages themselves. This work is thus an instance of the wider framework of Emergent Communication Fine-tuning (EC-FT) \cite{shane_etal_2022}.

In what remains, we describe our pipeline for EC fine-tuning (Section~\ref{sec:methods}) and the experiments that we conduct to demonstrate its benefit for UNMT (Section~\ref{sec:methods_experiments}), overview our experimental results, in which we show EC yields benefits for every language we study with particularly strong gains for the low-resource language Nepali (Section~\ref{sec:results}). We then study some manipulations on our training pipeline (Section~\ref{sec:manipulations}) before discussing the implications of these experiments (Section~\ref{sec:discussion}), and situating them in the context of existing work (Section~\ref{sec:related}).

Our contributions are the following:
    (i) We demonstrate that EC-FT can be used to improve upon UNMT baselines.
    (ii) We give a proof-of-concept for the viability of using modern pre-trained language models in an EC scenario.
    (iii) We articulate a view for EC-FT as a generalized and parameterizable framework.

\section{Methodology}
\label{sec:methods}
\begin{figure*}[ht]
  \centering
  \includegraphics[width=\textwidth]{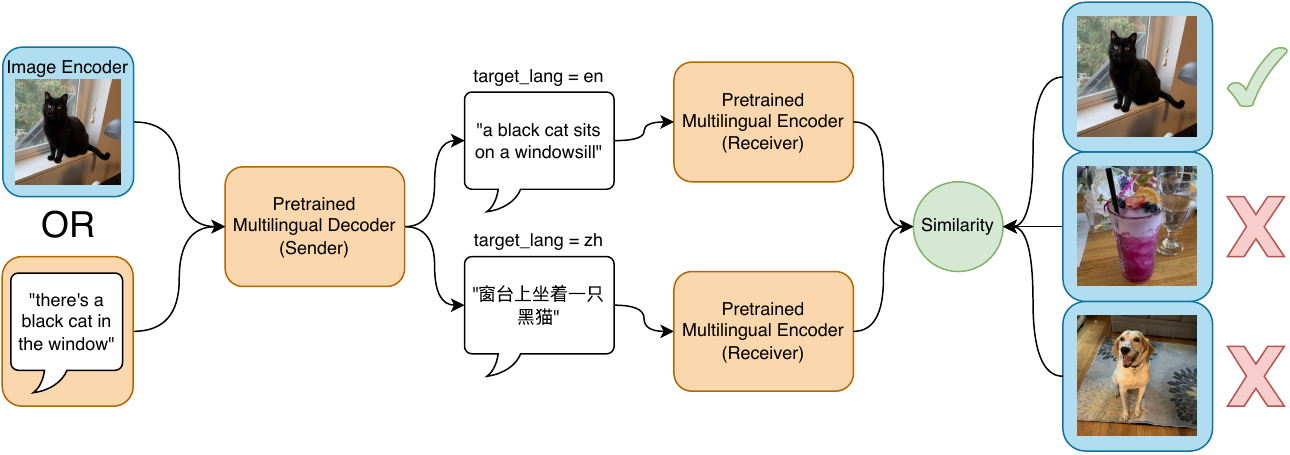}
  \caption{Emergent Communication Fine-Tuning: the task is a standard image reference game from the EC literature, but with the sender and receiver initialized from a pre-trained multilingual decoder and encoder. The communication language alternates between the two languages in the translation pair that is being fine-tuned.}
  \label{fig:model}
\end{figure*}

As shown in Figure~\ref{fig:introFig}, the pipeline that we introduce here consists of three main phases: (1) Begin with a pre-trained multilingual model, which either already has an encoder and decoder, or from which this \textit{seq2seq} stack can be initialized. (2) Conduct emergent-communication training using image and/or text embeddings (Figure~\ref{fig:model}). (3) Use iterative backtranslation (\citealp{sennrich-etal-2016-improving}; Section~\ref{sec:related}) to tune the model for translation.\footnote{The code to run all experiments described here can be found at \url{https://github.com/CLMBRs/communication-translation}.}

For step (2), we test two versions of the EC fine-tuning task. In the first ({I2I-EC}), the EC step uses \textit{only image embeddings}, and the model must select the original input image from among distractors, based on a text generation (akin to a caption). In the second ({T2I-EC}), the communication game involves gold captions, instead of only image features: based on a caption, the model must generate a translation of it, on the basis of which the original image must be selected from amongst distractors.

First, we introduce some notation. We use $E_m$ and $D_m$ for the multilingual encoder and decoder, respectively, which are parameterized by $\theta_E$ and $\theta_D$. $\textbf{x}_E \in \mathbb{R}^{N \times |V|}$ and $\textbf{x}_D \in \mathbb{R}^{K \times |V|}$ are sequences of symbols of length $N$ and $K$ respectively. $E_m(\textbf{x}_E ; \theta_E) \in \mathbb{R}^{N \times d_m}$, where $d_m$ is the model hidden dimension, is the encoder output. Similarly, the decoder output is $D_m(\textbf{x}_D, \textbf{e} ; \theta_D) \in \mathbb{R}^{K \times |V|}$, where $\textbf{e} \in \mathbb{R}^{N \times d_m}$ is a set of vectors for cross-attention of the decoder.

This formulation of our pipeline leaves many concrete choices open. In the remainder of this section, we describe the specific implementation of this process used in our experiments.

\subsection{Pipeline Components}
\label{sec:methods_pipeline}

\paragraph{Pre-trained Model}
We use mBART(-large) \citep{liu-etal-2020-multilingual-denoising}, which has demonstrated strong unsupervised translation performance in several languages. mBART employs \textit{seq2seq} pre-training, encoding a ``noised'' input sequence and then reconstructing the original sequence with the decoder, over a collection of 25 languages.
mBART's encoder-decoder architecture and corresponding seq2seq training make it a natural fit for our EC experiments, in which a multilingual decoder and encoder are used to send and receive natural language messages.
We use $\theta_{E_{PT}}$ to denote the parameters of the pretrained encoder, and \emph{mutatis mutandis} for the pretrained decoder.

\paragraph{Backtranslation}
Iterative backtranslation allows a model (usually pre-trained) to achieve some level of translation performance while only training on monolingual data (Section~\ref{sec:related}). Our baseline system is mBART fine-tuned with backtranslation only. In the EC-FT case, backtranslation is always performed last so that the model is tuned for translation immediately before it is evaluated.

\paragraph{Image-to-Image EC (I2I-EC)}
Our emergent communication framework consists of two main subtasks. First, an agent (the sender, a decoder) must take in an image encoding and produce a natural language description of it. The generation language may vary; there will be several in our experiments. Next, another agent (the receiver, an encoder) takes in the generated text and uses it to pick the described image from a set of distractors. In the EC literature, this is referred to as a standard image reference game (see Figure~\ref{fig:model}).\footnote{The image reference game, as used in much of the EC literature, is very similar to? training an image caption module to produce discriminative captions via self-retrieval, as pursued in \cite{liu_show_2018}. They first train the text-to-image pipeline from gold captions, and then pursue training a caption generator via image selection both with and without supervision from gold captions. We thank an anonymous reviewer for calling our attention to this work.}

Let $i \in \mathbb{R}^{d_i}$ be an image embedding ($d_i$ is the dimension of these embeddings, which may come from a vision model).  We also assume that we have a reshaper $R(i ; \theta_R)$ which maps images to $\mathbb{R}^{d_m}$.  

Because mBART is not natively multi-modal, some adaptations are made to allow it to generate a description of an image. In particular, the image embedding cannot simply be the first token to the sender since mBART reserves this for a special language identification token. Further, it is not obvious that a pre-trained transformer decoder's cross-attention can be ``turned off'' without effecting overall performance. For these reasons, we pass the image embedding into an ``unroller'' $U$ (one auto-regressive transformer layer) to generate a sequence of embeddings $U(R(i); \theta_U) \in \mathbb{R}^{M \times d_m}$ where $M$ is a hyperparameter. This sequence is then used as the keys and values in the sender's cross-attention.

We auto-regressively generate from the sender's distributions $S := D_m\left(\langle \text{LID}, T_{<K}\rangle , U(R(i))\right) \in \mathbb{R}^{K \times |V|}$, where LID is a language ID token and $T_{<K}$ is the prefix of text $T$ generated at the previous time step.
The sampling required for discrete generation is not differentiable, so we use the straight-through Gumbel-Softmax estimator \citep{jang_categorical_2017, maddison_concrete_2017} with temperature $\tau = 1.0$.  $T := \text{GS-ST}(S)$ is the sequence of one-hot vectors sampled in this way.

The receiver consumes this generated `caption': $E_m(T) \in \mathbb{R}^{K \times d_m}$.  To produce a single representation of the image, we use an `aggregator' $A$ which takes this sequence of representations and pools them into a single one $A(E_m(T); \theta_A) \in \mathbb{R}^{d_m}$.\footnote{Pilot experiments suggested that a small aggregator worked better than simply using mean pooling.}

The score for each of the candidate images is the inverse of the mean squared error between the image and the receiver's final representation. The loss for the image selection task is then cross-entropy among the image candidates. This loss partially follows \citet{lee_emergent_2018}, though they jointly train on supervised caption generation during EC.

Given the original image $i$, and a set $\{ i_m \}_{m=1}^M$ of distractor images, let the image selection loss be
\begin{align}
     \label{eqn:image-selection-loss}
& \ell_{\text{IS}} \left(i, \Theta\right) := \nonumber \\
&-\log \text{softmax} \frac{1}{\lVert A(E_m(T)) - R(i) \rVert_2^2}
\end{align}
where $\Theta = \{ \theta_D, \theta_E, \theta_R, \theta_A, \theta_U \}$ and the softmax is taken over the distractor images $\{R(i_m)\}$.

Finally, because EC can cause significant language drift \citep{lee_emergent_2018, lee-etal-2019-countering, lu_countering_2020, lazaridou-etal-2020-multi}, we use KL regularization \citep{havrylov_emergence_2017, baziotis-etal-2020-language} to ensure that the sender's output distribution does not drift too far from the distribution of an auxiliary causal language model (CLM; this model is not trained as part of EC):\footnote{We finetune the original mBART decoder as a CLM for this purpose; see the end of Appendix~\ref{appendix:main_params} for details.}
\begin{equation}\label{eqn:kl-reg}
\ell_{\text{KL}} := \frac{1}{K} \sum_k \text{KL}\left( S_k \mid\mid D_{\text{CLM}}\left( \langle LID, T_{<k} \rangle \right)_k \right)
\end{equation}

Combining equations~(\ref{eqn:image-selection-loss}) and~(\ref{eqn:kl-reg}) and averaging over iterations of the game, the final EC loss is
\begin{equation}\label{eqn:ec-loss}
\mathcal{L}_\text{EC} := \mathbb{E}_{i} \left[ \ell_\text{IS} + \lambda \ell_\text{KL} \right]
\end{equation}
with $\lambda$ a hyperparameter.

\paragraph{Text-to-Image EC (T2I-EC)}
The text-to-image EC task is identical to I2I-EC, except in what is presented to the sender via cross-attention. In T2I-EC, monolingual gold captions are used in the cross-attention for the emergent generation after being embedded by the encoder $E_m$.

In other words, given $c_i$ as a caption for image $i$, T2I-EC still uses $\mathcal{L}_\text{EC}$ (equation~(\ref{eqn:ec-loss})), but without the unroller for the sender.  Now, we have $S = D_m\left(\langle \text{LID}, S_{<K} \rangle , E_m(c_i ; \theta_{E}) \right)$.

As in I2I-EC, the image descriptions are generated in \textit{either} the caption language (here, English) or another translation target language. Importantly, the emergent generation need not be identical to the gold caption. This is desirable, since there may be several valid paraphrases of a given translation/caption.  Similarly, we only require gold captions in one language, not every language; for this reason, there is no implicitly parallel text data and so the translation task can still be considered unsupervised.

The motivation for this version of EC comes from the observation that the encodings used in the sender's cross-attention should be fairly similar to those generated by the model's encoder, since the model is being fine-tuned to be an encoder-decoder translation model. Generating into varying target languages incentivizes the model to use the same encodings for generating different languages, rather than copying the input text to the output. In contrast, there is no guarantee that the image encodings used in I2I-EC are at all similar to those produced by the model's encoder.

\paragraph{Initial Supervision}
Because multilingual EC is a complicated task with sparse training signal, we first ground the agents in their visual sub-tasks independently of the combined communication task. We train the sender to produce gold-standard captions in a high-resource language (English in our experiments) while simultaneously training the receiver to pick out the correct image based on the gold-standard caption. Critically, this stage only assumes that you have gold-standard captions in \textit{one} language. The model is never trained on gold captions in non-English languages. This step is conducted independently, before EC.

\subsection{Data}
\label{sec:methods_data}
\paragraph{Training}
We use two main sources of training data: monolingual corpora for backtranslation, and pairings of images and captions in a single high-resource language. We train translation systems between English and four other languages: Chinese (zh), German (de), Nepali (ne), and Sinhala (si).

Backtranslation creates synthetic translation pairs by generating sentences in the second language given natural sentences in the first. Following experiments using mBART for unsupervised translation \citep{liu-etal-2020-multilingual-denoising}, we use small portions of the Common Crawl 25 dataset, which is the pre-training data for mBART. In this way, no novel data is introduced to establish our UNMT baseline.

For the EC stage, the data required differs between I2I-EC and T2I-EC. The former requires only image embeddings. The latter requires paired images and captions, since the true caption is used to prompt the sender's generation. As mentioned, we assume that captions are \textit{only available for one language}. Since English is in every translation pair, we use English captions. Our image-caption pairs come from the MS-COCO dataset \citep{lin_microsoft_2014}, and our image embeddings are extracted from ResNet 50 \citep{he_deep_2016} (these are also used during the supervised captioning stage).

\paragraph{Validation and Test}
Translation validation and test sets are the only parallel data used in our experiments. For Nepali and Sinhala, we use the standard splits of the FLoRes evaluation datasets \citep{guzman-etal-2019-flores}. For Chinese and German, we use the \texttt{newstest2018} and \texttt{newsdev2019} splits of the WMT'19 release as validation data \citep{barrault-etal-2019-findings}. For test data in these two languages, we sample 4096 examples from News Commentary v14 subset of the same release.

\section{Experiments}
\label{sec:methods_experiments}
We evaluate a UNMT baseline and our two proposed EC-FT pipelines on translation performance for each language pair. Checkpoints are picked by highest mean BLEU on the validation set. We first describe these models and then our evaluation. More extensive details can be found in Appendix~\ref{appendix:main_params}.

\paragraph{Baseline}
For our UNMT baseline, we start with mBART-25 and perform iterative backtranslation for 8192 steps in each direction.  mBART employs language control tokens at the beginning of sequences, but it is \textit{not} pre-trained to decode one language from another \citep{liu-etal-2020-multilingual-denoising}, which is a key feature of (back-)translation. To overcome the model's tendency to copy the input sequence to the output, we establish language-controlled generation using language control tokens and language masks \citep{liu-etal-2020-multilingual-denoising}. Concretely, we obtain token counts from the mBART training data, and these are used to create a logit mask, only allowing the model to generate tokens which make up the top $p$ percent of the probability mass of the data in the given language. For the first 2048 backtranslation steps, we use a masking threshold of $p=0.9$. After that, we raise the threshold to $p=0.99$.

\paragraph{(I2I/T2I)-EC}
In both of our EC-FT models, we keep the total number of backtranslation steps the same (8192), and add 2048 steps each of supervised caption training and EC-FT. The language of generation can also be controlled during EC, so we use language-control tokens and a logit mask to ensure the sender generates in the specified language. The language of the emergent generation is selected uniformly at random per example.

\paragraph{Evaluation} For our final evaluation, we report both BLEU and COMET \citep{rei-etal-2020-comet} scores in both translation directions for each language pair. COMET provides the output of a regression model trained to predict the human direct-assessment translation quality score of a translation pair. Based on normalized quality scores, a COMET score of 0 means the translation is predicted to be of average quality. Postive scores indicate above-average quality, and vice-versa.  We use the \texttt{wmt22-comet-da} model.

\begin{table*}[ht!]
  \centering
  \begin{tabular}{lccccccccc}
    \toprule
    \multirow{2}{*}{Model} & \multirow{2}{*}{Language} & \multicolumn{4}{c}{BLEU} & \multicolumn{3}{c}{COMET}
    \\ \cmidrule(lr){3-6} \cmidrule(lr){7-9}
    {} & {} & en$\to$X & X$\to$en & mean & $\Delta$ & en$\to$X & X$\to$en & mean 
    \\
    \midrule
    \multirow{4}{*}{baseline (mBART + BT)} & zh & 18.45 & 11.36 & 14.90 & - & 0.03 & 0.15 & 0.09 
    \\
    {} & de & 19.06 & 25.73 & 22.39 & - & 0.20 & 0.38 & 0.29 
    \\
    {} & ne & 2.14 & 5.07 & 3.60 & - & -0.24 & -0.34 & -0.29
    \\
    {} & si & 1.18 & 4.73 & 2.95 & - & -0.18 & -0.28 & -0.23
    \\
    \midrule[0.8\lightrulewidth]
    \multirow{4}{*}{I2I-EC} & zh & 18.72 & 11.88 & 15.30 & +3\% & 0.04 & 0.17 & 0.10 
    \\
    {} & de & 18.26 & 25.60 & 21.93 & -2\% & 0.20 & 0.40 & \textbf{0.30} 
    \\
    {} & ne & 1.51 & 5.34 & 3.43 & -5\% & -0.24 & -0.31 & -0.28
    \\
    {} & si & 0.01 & 0.08 & 0.04 & -99\% & -1.31 & -1.05 & -1.28
    \\
    \midrule[0.8\lightrulewidth]
    \multirow{4}{*}{T2I-EC} & zh & 19.25 & 11.91 & \textbf{15.58} & +5\% & 0.06 & 0.18 & \textbf{0.12}
    \\
    {} & de & 18.64 & 26.20 & \textbf{22.42} & +0.1\% & 0.19 & 0.41 & \textbf{0.30}
    \\
    {} & ne & 2.36 & 5.92 & \textbf{4.14} & +15\% & -0.20 & -0.27 & \textbf{-0.24}
    \\
    {} & si & 1.29 & 4.76 & \textbf{3.02} & +2\% & -0.18 & -0.27 & \textbf{-0.22}
    \\
    \bottomrule
  \end{tabular}
  \caption{Results of our main experiment.  Values reported here are the maximum across 3 random seeds per row; see Appendix~\ref{appendix:full-results} for full variation.  T2I-EC shows consistent improvement for each language in terms of both mean BLEU and COMET. $\Delta$ shows percent improvement over the baseline.}
  \label{tab:main-results}
\end{table*}

\section{Results}
\label{sec:results}

Table~\ref{tab:main-results} shows the results from our main experiment.
Firstly, our UNMT baseline based on iterative backtranslation (BT) shows a marked decrease in performance from the two higher-resource languages (Chinese and German) to the two lower-resource languages (Nepali and Sinhala). This is expected since BT-based UNMT often requires a strong  initialization \citep{lample-etal-2018-phrase} and multilingual models (like mBART) do not perform as well for lower-resource languages \citep{wu-dredze-2020-languages}.

Our model fine-tuned with both backtranslation and I2I-EC remains close to or exceeds the baseline for the two higher-resource languages and Nepali but achieves very poor performance on Sinhala. It appears that EC provides a worse initialization for backtranslation for this language.

In contrast,  our ``text-to-image'' variant of EC-FT (T2I-EC) yields the best performing model in terms of mean BLEU for all four of our languages. In particular, we see significant gains for both lower-resource languages. Most striking is the Nepali-English pair, which sees a +15\% BLEU improvement over the baseline.  While there are improvements in both directions, the Nepali$\to$English direction has the largest gain.  By contrast, Sinhala shows improvements in both directions, with the larger improvement in the to-Sinhala direction (partially due to a stronger baseline). The improvements are smallest for German, which is both very high-resource and the most similar to English of our languages.
The COMET scores were broadly correlated with BLEU scores in all of our settings.

These results show that EC-FT of a pre-trained multilingual model can provide real improvement over a backtranslation-only baseline, giving proof-of-concept of communication for fine-tuning.

\section{Manipulations} 
\label{sec:manipulations}
To better understand which components of the pipeline affect the results in T2I-EC, we conducted several follow-up experiments.  For each manipulation, we looked at one high-resource language (German) and one low-resource langauge (Sinhala). See Appendix~\ref{appendix:manipulations_params} for full methodological details.

\paragraph{Image Encoder} To test the effect of the image encoder, we replaced the ResNet image encoder with the best performing one from CLIP \citep{radford_learning_2021}.  This image encoder is based on the Vision Transformer \citep{dosovitskiy_image_2021} architecture and trained jointly with a text encoder via a contrastive loss to pair image encodings with caption encodings.

\paragraph{Initial Backtranslation} Because the EC component of training is the first time that language ID codes are being used to generate text from the decoder with input other than representations of the same language from the encoder, we experimented with splitting the backtranslation training into two parts.  Instead of doing all 8192 steps after EC, we did 2048 steps after image supervision but before EC, and the final 6144 steps after EC.

\paragraph{Interleaved Training} Inspired by \citet{loweInteractionSupervisionSelfplay2020}, who showed that inter-leaving EC with a supervised learning objective can improve EC results, we ran a version of our training pipeline where we alternated between EC and BT four times.  The total number of training steps remained the same (2048 and 8192, respectively), but this was now done in 4 equal-sized EC-to-BT pieces.

\paragraph{Results}  Table~\ref{tab:ablation-results} shows the results of these ablations.  Evaluation is in terms of BLEU on the test set, and the $\Delta$ column reports the percent difference from the best value for a language in Table~\ref{tab:main-results}. We find significant reduction in translation quality with the CLIP image encoder and inconsistent performance
for both an initial BT phase and interleaved training, with performance dropping for German but slightly increasing for Sinhala when compared to T2I-EC (as seen in the $\Delta$ column).

\begin{table}[ht!]
\resizebox{\columnwidth}{!}{%
    \begin{tabular}{lccccc}
    \toprule
    Manipulation & Lang & en$\to$X & X$\to$en & mean & $\Delta$
    \\
    \midrule
    CLIP-img & de & 18.52 & 25.93 & 22.23 & -1\%
    \\
    CLIP-img & si & 1.05 & 4.18 & 2.61 & -14\%
    \\
    \midrule
    Init BT & de & 18.20 & 25.39 & 21.80 & -3\%
    \\
    Init BT & si & 1.24 & 4.84 & 3.04 & +0.6\%
    \\
    \midrule
    Interleave & de & 18.29 & 25.69 & 21.99 & -2\%
    \\
    Interleave & si & 1.25 & 4.84 & 3.05 & +1\%
    \\
    \bottomrule
    \end{tabular}
}
\caption{Results from several training pipeline manipulations.  BLEU scores reported; $\Delta$ is the percentage difference from the corresponding mean value in T2I-EC in Figure~\ref{tab:main-results}.}
\label{tab:ablation-results}
\end{table}

\section{Discussion}
\label{sec:discussion}
We have demonstrated that (at least one variant of) EC fine-tuning provides improvement on unsupervised translation over a standard backtranslation baseline. The gains are especially pronounced for the low-resource language Nepali, which is ideal since under-resourced languages constitute the expected use case for unsupervised translation techniques. Furthermore, since the hyperparameters for the EC-FT portion of our pipeline were mostly determined empirically, our approach may be \textit{under-optimized}, meaning future work may yield further improvement using the same technique. 

\paragraph{I2I-EC} However, it is also clear that our formulation and implementation of ``standard'' EC (I2I-EC) does not improve upon the baseline, and even degrades performance in many cases. Our interpretation of this behavior is linked to our motivation for formulating T2I-EC in the first place. 

As mentioned in Section~\ref{sec:methods_pipeline}, the image representations used in the sender's cross-attention, in the image-to-image setup, are not guaranteed to be at all similar to the representations that the receiver learns to encode. Because we seek to fine-tune for a standard \textit{seq2seq} task (translation), it is desirable that the sender (mBART decoder) be trained to use the same or similar representations to those produced by the receiver (mBART encoder). Thus, we hypothesize that the null and negative effects of I2I-EC may be due to this mismatch between the representations the sender is trained to use, and those that the receiver is trained to produce.

However, we do \textbf{not} believe we have shown that I2I-EC will not be useful under slightly different formulations. In particular, the image representations may be able to be constrained to be similar to those of the receiver, either during EC or during the initial supervision phase. This could be accomplished using an auxillary distance loss, or by normalizing the mean and variance of the representations in both places.

\paragraph{EC Fine-Tuning}
Lastly, we view EC fine-tuning as a broader framework in which we have tested two distinct formulations \citep{shane_etal_2022}. We will assume that the invariant element of EC is a model's use of discrete, natural-language generations as input to a second model, which must use them to accomplish some task.

Given this definition, there are several choice points for applying EC-FT. The parameter we explicitly explore in our experiments is whether the input to the sender is \textit{image-based} or \textit{text-based}. In both of our formulations, the receiver is trained by a contrastive image-choice loss. Another parameter for future work concerns whether this loss applies to images or texts. The receiver could be trained to choose the correct sentence out of a set of distractors via the similarity of the sentence embeddings.

 A third parameter is whether the receiver is trained by a \textit{contrastive} loss or a \textit{generative} one (i.e. exactly reproducing a target sequence, as in \textit{seq2seq} training).\footnote{Known as ``reference game'' versus ``reconstruction game'' in the EC literature \citep{lazaridou_emergent_2020}.} In fact, an EC parameterization with text input, text output, and generative loss has already been formulated elsewhere, though it is not referred to as such. \citet{niu-etal-2019-bi} design a formulation of backtranslation, in which the artificial intermediate text is generated with straight-through Gumbel Softmax, instead of generated separately first. Future work will explore using this method with pre-trained models, i.e.\ in an EC-FT context.

These and other parameter choices leave extensive room for exciting future work with EC-FT as a general framework, both for UNMT and beyond.

\section{Related Work}
\label{sec:related}

\paragraph{UNMT}
\label{sec:unmt}
Unsupervised NMT uses only monolingual texts in each language of interest. \citet{lample-etal-2018-phrase} describe three principles for successful UNMT systems: 1. \textit{initialization}, the initial model must leverage aligned representations between languages; 2. \textit{language modeling}, there should be a strong ``data driven prior'' over the text patterns of each language; and 3. \textit{backtranslation} which turns the unsupervised problem into a noisily-supervised one, through the use of semi-synthetic translations.

Significant progress has been made in improving each of these aspects of UNMT. Pre-trained multilingual language models \citep{lample_cross-lingual_2019, conneau-etal-2020-unsupervised,liu-etal-2020-multilingual-denoising,song_mass_2019} have vastly improved the tractability of principles 1 and 2, largely replacing initialization techniques using inferred bilingual dictionaries \citep[e.g.][]{lample_2018_word}.

For the third principle, \textit{iterative backtranslation} is widely used \citep{sennrich-etal-2016-improving,he_dual_2016,lample_unsupervised_2018,haddow_survey_2022}. On this approach, synthetic data is generated ``on the fly'', during training. The model is updated before each new batch of synthetic text is generated, leading to simultaneous incremental improvement in generated data quality and model quality.

In this work, we adhere to all three principles, but add EC as a training signal. 
It has been noted that UNMT baselines still perform relatively poorly for low-resource languages \citep{guzman-etal-2019-flores}. We improve upon low-resource UNMT pipelines by leveraging goal-directed, multimodal fine-tuning via emergent communication.

\paragraph{EC and NLP}
\label{sec:ec-nlp}
A few other papers combine EC and NMT specifically. 
\citet{lee_emergent_2018} use EC and image captioning to build UNMT models, showing that EC promotes better translation than the multimodal alignment technique of \citet{nakayama_zero-resource_2017}.  Our approach differs in several important respects: we initialize our EC environment with \emph{pre-trained language models}; we use both EC and backtranslation; and we do not simultaneously train on the EC objective and image captioning objective. Moreover, because we use one multilingual model, our caption grounding only uses one language, instead of all languages. Our results show that EC promotes unsupervised translation in the context of advanced methods that combine pre-training with backtranslation.

\citet{li-etal-2020-emergent} use emergent communication as a pre-training step for NMT systems.  They have agents play an EC game, and then use those parameters to initialize an NMT system.  They find that (together with adapters and weight-distance regularization) EC pre-training improves in BLEU over a standard NMT baseline, with especially large gains coming in the few-shot setting. While this shows that EC can provide a good initialization for a recurrent NMT system, our present work shows that EC can provide a good fine-tuning signal for a pre-trained multilingual language model. We also note two differences with respect to both works: (i) they use recurrent networks, whereas we start from a pre-trained transformer, and (ii) they use separate models for each language, whereas we use one multilingual model. 

\citet{lee-etal-2019-countering} cast translation as a communication game with a third pivot language as the latent space in order to study (i) language drift from a pre-trained supervised MT model and (ii) using visual grounding (via gold image captions) plus language modeling to counter such drift. This approach thus does use EC with a pre-trained model, but it is a small model trained on the target task (translation). Our approach encourages using EC in conjunction with large-scale pre-trained language models which are intended to be general-purpose.
 
Finally, \citet{lazaridou-etal-2020-multi} study various ways of combining EC with a standard vision-language task, namely image captioning. They identify several forms of language drift and explore ways of incorporating auxillary losses. This work heavily inspires our own, since many of their settings correspond to using a pre-trained image-caption system. Our focus, however, has been on using EC to fine-tune large-scale pre-trained models on a language-only task, which introduces its own challenges and has its own benefits.

\paragraph{Multimodal pre-training}
\label{sec:multimodal}
Recently, efforts in multimodal pre-training are surging, especially in vision-language (V-L) pre-training \citep{du_survey_2022}. Most of the works create joint V-L representations through a fusion encoder \citep{li_unicoder-vl_2019, li_visualbert_2019, tan-bansal-2019-lxmert}, where the fused representation is the joint representation of image and text, as learned by a single encoder. Other recent works such as CLIP \citep{radford_learning_2021} and ALIGN \citep{jia_scaling_2021} attempt to use different encoders for images and text to make the framework more efficient. While V-L pre-training models image and text data jointly \citep{du_survey_2022, wang_ufo_2021}, we start with an existing pre-trained language model and further train it through the communication process in an image referential game. Although we expect the alignment between image and text to arise through this process, we view the visual modality as an additional signal to ground the multilingual communication process.

We also note that most previous work on V-L pre-training is evaluated solely on vision or V-L tasks \citep{li_visualbert_2019, radford_learning_2021, jia_scaling_2021}. The advantage of this joint pre-training for language-only tasks remains unclear \citep{yun_vision-and-language_2021, pezzelle_word_2021}. In this paper, we focus on a language-only task (UNMT) to evaluate whether visual grounding can improve such tasks.

Finally, we note that EC-FT is more general than typical approaches to multimodal pre-training. While the image-based task we employ here works by promoting multimodal alignment, the range of possible tasks that can be used in EC-FT is huge, from directing other agents \citep{mordatch_emergence_2018} to controlling a robot \citep{das_tarmac_2019} to playing games and reasoning about social dilemmas \citep{jaques_social_2019}. This wide range of tasks can incorporate many dimensions of communication that should be beneficial for NLP systems---e.g.\ other agents with their own private goals, social context, embodied control---that are not easily captured by multimodal pre-training \citep{bender-koller-2020-climbing, bisk-etal-2020-experience}. In terms of \citet{bisk-etal-2020-experience}'s \emph{world scopes} mentioned in the introduction, multimodal pre-training corresponds to world scope 3 (perception); EC-FT has the ability to move us much closer towards the final scopes 4 (embodiment and action) and 5 (the social world). 

\paragraph{Multimodal Fine-tuning} A related body of work focuses on adapting pre-trained language-only models for use in multi-modal tasks. For example, \citet{tsimpoukelli_multimodal_2021} show that using a frozen language model and adapting a visual encoder to produce embeddings aligned with the LM's can be useful for few-shot learning in multimodal tasks like visual question answering. \citet{liang_modular_2022} make this approach more modular by additionally freezing the visual encoder and learning separate prompt vectors. In the EC-FT context, these works suffer some of the same limitations in world scope, but could provide very useful methods for the environment-to-sender adapter step discussed in Section~\ref{sec:methods_pipeline}. 

\section{Conclusion}
\label{sec:conclusion}
We have shown that Emergent Communication can be used as a fine-tuning signal for a large pre-trained multilingual model; this grounding in a goal-oriented multimodal task yields improvements over an unsupervised NMT baseline in all four languages studied.
There is likely room to further improve upon the specific EC variants we propose here, since we believe the EC process is under-optimized for hyperparameters. We have further noted that the framework we propose leaves extensive room for further experimentation, since there are many choice points of the general EC setup that we have not yet tested, and may be promising avenues for future improvement.  The general EC-FT framework may also be applied to other tasks beyond UNMT in future work.

\section*{Author Contribution Statement}

Following a practice in several other fields, we here list author contributions according to the Contributor Role Taxonomy (CRediT; \citealp{allenHowCanWe2019}).
\textbf{C.M. Downey:} Conceptualization, Methodology, Software, Validation, Investigation, Writing - original draft, Writing - review and editing, Visualization. \textbf{Xuhui Zhou:} Conceptualization, Methodology, Software, Validation, Investigation, Data curation, Writing - review and editing. \textbf{Zeyu Liu:} Conceptualization, Methodology, Software, Validation, Investigation, Data curation, Writing - review and editing. \textbf{Shane Steinert-Threkeld:} Conceptualization, Methodology, Resources, Writing - original draft, Writing - review and editing, Supervision, Project administration, Funding acquisition.

\section*{Limitations}
One limitation of our work concerns analysis. Much remains to be learned about the mechanisms by which EC can help translation. By evaluating the model more comprehensively, we could gain insight into whether and how the grounding helps task performance. Based on such analysis, a better version of the pipeline could be developed. 

We observed significant variability across random seeds in our EC training; methods for stabilizing this variability could ensure the reliability of EC as a fine-tuning process for models.

Finally, we investigated only four non-English languages, two `high-resource' and two `low-resource'.  It would be valuable to explore a wider range of typologically diverse languages to validate that these methods apply across the board and, if not, to understand what language factors drive success.

\section*{Ethics Statement}
This work on unsupervised translation should have a positive impact on many under-served language communities by extending the reach of a core language technology (translation) to languages which lack the extensive parallel data required for supervised translation systems.

That being said, there are ethical risks with the present approach. The pre-training of mBART depends on the CommonCrawl dataset, so there might be some offensive language and even identity leakage due to CommonCrawl's preprocessing pipeline. It is possible that the model will generate toxic and biased utterances in our experiments. We didn't evaluate the toxicity of our generation. Our intuition is that the caption grounding will bias the model towards descriptive captions and thus suppress the toxic generation.

\subsubsection*{Acknowledgments}

We thank Emily M Bender, Emmanuel Chemla, Chris Potts, Tania Rojas-Esponda, and the anonymous reviewers of and audience at the ICLR 2022 Emergent Communication workshop for helpful discussion. This work was partially supported by funding from the University of Washington Royalty Research Fund (RRF), grant number A167354 ``Learning to translate by learning to communicate''.

\bibliographystyle{acl_natbib}
\bibliography{tmlr/anthology,tmlr/ec-ft}

\appendix
\section{Main Experiments Training Details}
\label{appendix:main_params}
We here include more details about the training protocol for the results reported in Section~\ref{sec:results}. Our codebase is built upon the mBART code from \texttt{huggingface} \citep{huggingface} and PyTorch \citep{pytorch}. We use one NVIDIA RTX 8000 GPU for each experiment. Backtranslation is the most expensive part of the entire training pipeline. It takes around 24-28 hours to finish, depending on the languages. The combined training time for caption grounding and emergent communication is within 1 hour.

\paragraph{Baseline}
As discussed in section \ref{sec:methods_experiments}, our UNMT baseline is established by starting with mBART and performing 8192 steps of iterative backtranslation for each translation pair. We use a batch size of 32 and a maximum generated sequence length of 64. See more hyperparameter choices in Table~\ref{tab:bt-hyper}.

\paragraph{I2I-EC} For our I2I-EC fine-tuned model, training consists of the following pipeline
\begin{enumerate}
  \item 2048 steps of backtranslation
  \item 2048 steps of supervised captioning training (English-only)
  \item 2048 steps of EC fine-tuning
  \item 6144 steps of backtranslation
\end{enumerate}

Backtranslation uses the same exact hyperparameters as in the baseline, but with training split between the first 2048 and last 6144 steps (Table~\ref{tab:bt-hyper}).

Supervised caption training is described in Section~\ref{sec:methods_pipeline}. We have 8 choices for the image selection task (7 distractors and 1 correct choice). As part of Sender agent, we use a one-layer auto-regressive transformer to serialize (or, ``unroll'') a single ResNet image representation to a sequence of vectors to imitate the sequential data mBART observes during its pre-training. The unrolled sequence is used in the sender's cross-attention, and the sender is trained to generate the gold-standard caption.

Also during the supervised captioning stage, the receiver takes in the gold-standard caption, and a one-layer RNN is used to aggregate its final hidden states and choose the correct image. The image selection (cross-entropy) loss is scaled with $\lambda$, before being added to the caption-generation loss. Full hyperparameter choices are detailed in Table~\ref{tab:captioning-hyper}.

I2I-EC fine-tuning is also described in Section~\ref{sec:methods_pipeline}. Different from caption grounding, we have a total of 16 image choices instead of 8. 
The adapter unrolls the ResNet image representation to a length of 32. The emergent generation is language-constrained as described in Section~\ref{sec:methods_experiments} with a threshold. A repetition penalty is applied to the generations, and they are constrained to not repeat any 4-grams or longer. KL-regularization with a separate mBART instance fine-tuned on causal language modeling is applied with a $\lambda$ parameter. Full hyperparameter choices are detailed in Table~\ref{tab:ec-hyper}.

\paragraph{T2I-EC} For our T2I-EC fine-tuned model, training is performed slightly differently for empirical reasons
\begin{enumerate}
  \item 2048 steps of supervised captioning training (English-only)
  \item 2048 steps of EC fine-tuning
  \item 8192 steps of backtranslation
\end{enumerate}

T2I-EC hyperparameters are very similar to I2I-EC. See full parameters in Tables~\ref{tab:bt-hyper}, \ref{tab:captioning-hyper}, and \ref{tab:ec-hyper}.

\begin{table*}[h] 
    \centering
    \begin{subtable}[h]{1\linewidth}
        \centering
        \begin{tabular}{ll}
            \toprule
            \textbf{Name} & \textbf{Values}  \\
            \toprule
                optimizer & \texttt{Adam(betas=(0.9, 0.999))} (default in PyTorch) \\
                LR scheduler & \texttt{constant\_w\_warmup} \\
                \texttt{grad\_clip} & 1.0 \\
                \texttt{batch\_size} & 32 \\
                \texttt{evaluate\_bleu\_every} & 256 \\
                $|$validation\_set$|$ & 4096 \\
                \#beams & 5 \\
                first \texttt{\#vocab\_constrained\_steps} & 2048 \\
                threshold (after \texttt{\#vocab\_constrained\_steps}) & 0.99 \\
                \#warmup\_steps & $\frac{1}{4} \cdot$ \#steps \\
            \bottomrule
        \end{tabular}
        \caption{Backtranslation shared parameters}
        \label{tab:ec-hyper-shared}
    \end{subtable} 
    \begin{subtable}[h]{0.45\linewidth}
        \centering
        \caption{Baseline}
        \label{tab:bt-hyper-baseline}
        \begin{tabular}{cc}
            \toprule
            \textbf{Name} & \textbf{Values}  \\
            \toprule
                Learning rate & 2.0e-5 \\
                \#steps & 8192 \\
                first\_threshold & 0.90 \\
            \bottomrule
        \end{tabular}
    \end{subtable} 
    \\
    \begin{subtable}[h]{0.45\linewidth}
        \centering
        \caption{I2I-EC (Initial BT)}
        \label{tab:bt-hyper-i2i-init}
        \begin{tabular}{cc}
            \toprule
            \textbf{Name} & \textbf{Values}  \\
            \toprule
                Learning rate & 1.0e-5 \\
                \#steps & 2048 \\
                first\_threshold & 0.96 \\
            \bottomrule
        \end{tabular}
    \end{subtable} 
    \hfill
    \begin{subtable}[h]{0.45\linewidth}
        \centering
        \caption{I2I-EC (Secondary BT)}
        \label{tab:bt-hyper-i2i-secondary}
        \begin{tabular}{cc}
            \toprule
            \textbf{Name} & \textbf{Values}  \\
            \toprule
                Learning rate & 1.0e-5 \\
                \#steps & 6144 \\
            \bottomrule
        \end{tabular}
    \end{subtable}\\
    \begin{subtable}[h]{0.45\linewidth}
        \centering
        \caption{T2I-EC}
        \label{tab:bt-hyper-t2i}
        \begin{tabular}{cc}
            \toprule
            \textbf{Name} & \textbf{Values}  \\
            \toprule
                Learning rate & 1.0e-5 \\
                \#steps & 8192 \\
                first\_threshold & 0.96 \\
            \bottomrule
        \end{tabular}
    \end{subtable}
    \caption{Hyper-parameters for backtranslation.}
    \label{tab:bt-hyper}
\end{table*}

\begin{table*}[h] 
    \centering
    \begin{subtable}[h]{1\linewidth}
        \centering
        \begin{tabular}{ll}
            \toprule
            \textbf{Name} & \textbf{Values}  \\
            \toprule
                optimizer & \texttt{Adam(betas=(0.9, 0.999))} (default in PyTorch) \\
                \#steps & 2048 \\
                learning rate & 4.0e-5 \\
                LR scheduler & \texttt{linear\_w\_warmup} \\
                \#warm-up steps & 0 \\
                batch\_size & 16 \\
                \#distractors & 7 \\
                Reshaper (Sender \& Receiver) & linear projection \\
                Dropout (anywhere) & 0.0\\
                Image Unroll & one (auto-regressive) transformer layer \\
                Image Unroll length & 32 \\
                Receiver aggregation & RNN \\
                Sender & no freezing \\
                Receiver & no freezing \\
                beam\_width & 1 (Greedy) \\
                temperature & 1.0 \\
                gumble\_softmax sample &  one-hot \\
                \texttt{repetition\_penalty} & 1.0 \\
                \texttt{max\_seq\_length} & 32\\
            \bottomrule
        \end{tabular}
        \caption{Captioning shared parameters}
        \label{tab:captioning-hyper-shared}
    \end{subtable} 
    \begin{subtable}[h]{0.45\linewidth}
        \centering
        \caption{I2I-EC}
        \label{tab:captioning-hyper-i2i}
        \begin{tabular}{cc}
            \toprule
            \textbf{Name} & \textbf{Values}  \\
            \toprule
                Image selection loss $\lambda$ & 4.0 \\
                grad\_clip & 1.0 \\
            \bottomrule
        \end{tabular}
    \end{subtable} 
    \hfill
    \begin{subtable}[h]{0.45\linewidth}
        \centering
        \caption{T2I-EC}
        \label{tab:captioning-hyper-t2i}
        \begin{tabular}{cc}
            \toprule
            \textbf{Name} & \textbf{Values}  \\
            \toprule
                Image selection loss $\lambda$ & 8.0 \\
                grad\_clip & 0.5 \\
            \bottomrule
        \end{tabular}
    \end{subtable}
    \caption{Hyper-parameters for caption grounding part of emergent communication.}
    \label{tab:captioning-hyper}
\end{table*}

\begin{table*}[h] 
    \centering
    \begin{subtable}[h]{1\linewidth}
        \centering
        \begin{tabular}{ll}
            \toprule
            \textbf{Name} & \textbf{Values}  \\
            \toprule
                optimizer & \texttt{Adam(betas=(0.9, 0.999))} (default in PyTorch)\\
                \#steps & 2048 \\
                LR scheduler & \texttt{linear\_w\_warmup} \\
                \#warm-up steps & 0 \\
                batch\_size & 12 \\
                \#distractors & 15 \\
                Reshaper (Sender \& Receiver) & linear projection \\
                Dropout (anywhere) & 0.0\\
                Image Unroll & one (auto-regressive) transformer layer \\
                Image Unroll length & 32 \\
                Receiver aggregation & RNN \\
                Sender & no freezing \\
                Receiver & no freezing \\
                beam\_width & 1 (Greedy) \\
                temperature & 1.0 \\
                gumble\_softmax sample &  one-hot \\
                \texttt{vocab\_constraint\_threshold} & 0.99 \\
                \texttt{repetition\_penalty} & 1.0 \\
                \texttt{max\_seq\_length} & 32\\
            \bottomrule
        \end{tabular}
        \caption{Emergent communication shared parameters}
        \label{tab:ec-hyper-shared}
    \end{subtable} 
    \begin{subtable}[h]{0.45\linewidth}
        \centering
        \caption{I2I-EC}
        \label{tab:ec-hyper-i2i}
        \begin{tabular}{cc}
            \toprule
            \textbf{Name} & \textbf{Values}  \\
            \toprule
                Language modeling loss $\lambda$ & 0.125 \\
                Learning rate & 6.0e-6 \\
                grad\_clip & 1.0 \\
            \bottomrule
        \end{tabular}
    \end{subtable} 
    \hfill
    \begin{subtable}[h]{0.45\linewidth}
        \centering
        \caption{T2I-EC. $^*$: length of text string in place of series of "pseudo-images" from image unroller}
        \label{tab:ec-hyper-t2i}
        \begin{tabular}{cc}
            \toprule
            \textbf{Name} & \textbf{Values}  \\
            \toprule
                Language modeling loss $\lambda$ & 0.0625 \\
                Learning rate & 1.0e-6 \\
                grad\_clip & 0.5 \\
                \texttt{max\_text\_seq\_length}$^*$ & 128 \\
            \bottomrule
        \end{tabular}
    \end{subtable}
    \caption{Hyper-parameters for emergent communication.}
    \label{tab:ec-hyper}
\end{table*}

\paragraph{Auxiliary CLM} To have a language model for use in KL regularization (see equation~(\ref{eqn:kl-reg})), we fine-tuned just the mBART decoder on the same common crawl data used for its pretraining in all of the languages of interest.  We trained for 100000 steps, batch size 32, sequence length 96, and learning rate of $6\times10^{-6}$.  This model was then frozen during EC training and only used to compute the KL divergence which was used in updating the sender's parameters.

\section{Manipulations Training Details}
\label{appendix:manipulations_params}
All manipulations are performed on the main T2I-EC process. Interleaved training uses versions of the the learning rate schedules used for the main experiments shortened by a factor of 4.

\section{Full results}
\label{appendix:full-results}
In Table \ref{tab:main-results-resnet-full}, we include full results for our main experiment (summarized in Table~\ref{tab:main-results}). Although we found the EC process to help with machine translation, it also leads to instability in model training. We a systematic study of this variation to future work. 

In Table~\ref{tab:main-results-clipl-full} we show experiments with a more modern choice of image encoder --- CLIP-Large \cite{mullenbach-etal-2021-clip}. We find that the CLIP-Large encoder under-performs ResNet.

The full results from our manipulation experiments (Section~\ref{sec:manipulations}) are found in Table~\ref{tab:ablation-results-full}.

\begin{table*}[ht!]
  \centering
  \begin{tabular}{lcccccccc}
    \toprule
    \multirow{2}{*}{Model} & \multirow{2}{*}{Language} & \multirow{2}{*}{Seed} & \multicolumn{3}{c}{BLEU} & \multicolumn{3}{c}{COMET}
    \\ \cmidrule(lr){4-6} \cmidrule(lr){7-9}
    {} & {} & {} & en$\to$X & X$\to$en & mean & en$\to$X & X$\to$en & mean
    \\
    \midrule
    \multirow{12}{*}{baseline (mBART + BT)} & \multirow{3}{*}{zh} & 1 & 17.21 & 11.35 & 14.28 & -0.04 & 0.14 & 0.05
    \\
    {} & {} & 2 & 18.38 & 11.39 & 14.89 & 0.02 & 0.14 & 0.08
    \\
    {} & {} & 3 & 18.45 & 11.36 & 14.90 & 0.03 & 0.15 & 0.09
    \\ \cmidrule(lr){2-9}
    {} & \multirow{3}{*}{de} & 1 & 18.66 & 25.83 & 22.24 & 0.18 & 0.39 & 0.29
    \\
    {} & {} & 2 & 19.06 & 25.73 & 22.39 & 0.20 & 0.38 & 0.29
    \\
    {} & {} & 3 & 18.79 & 25.88 & 22.33 & 0.22 & 0.40 & 0.31
    \\ \cmidrule(lr){2-9}
    {} & \multirow{3}{*}{ne} & 1 & 1.94 & 4.74 & 3.34 & -0.19 & -0.36 & -0.27
    \\
    {} & {} & 2 & 1.84 & 4.94 & 3.39 & -0.20 & -0.34 & -0.27
    \\
    {} & {} & 3 & 2.14 & 5.07 & 3.60 & -0.24 & -0.34 & -0.29
    \\ \cmidrule(lr){2-9}
    {} & \multirow{3}{*}{si} & 1 & 1.29 & 4.53 & 2.91 & -0.29 & -0.31 & -0.30
    \\
    {} & {} & 2 & 1.18 & 4.73 & 2.95 & -0.18 & -0.28 & -0.23
    \\
    {} & {} & 3 & 1.21 & 4.35 & 2.78 & -0.20 & -0.32 & -0.26
    \\
    \midrule[0.8\lightrulewidth]
    \multirow{12}{*}{I2I-EC} & \multirow{3}{*}{zh} & 1 & 17.31 & 10.96 & 14.13 & -0.03 & 0.12 & 0.05
    \\
    {} & {} & 2 & 17.03 & 11.24 & 14.14 & 0.00 & 0.15 & 0.07
    \\
    {} & {} & 3 & 18.72 & 11.88 & 15.30 & 0.04 & 0.17 & 0.10
    \\ \cmidrule(lr){2-9}
    {} & \multirow{3}{*}{de} & 1 & 18.22 & 25.41 & 21.81 & 0.18 & 0.39 & 0.29
    \\
    {} & {} & 2 & 18.26 & 25.60 & 21.93 & 0.18 & 0.39 & 0.29
    \\
    {} & {} & 3 & 18.06 & 25.28 & 21.67 & 0.20 & 0.40 & 0.30
    \\ \cmidrule(lr){2-9}
    {} & \multirow{3}{*}{ne} & 1 & 1.24 & 5.13 & 3.19 & -0.25 & -0.31 & -0.28
    \\
    {} & {} & 2 & 1.22 & 5.30 & 3.26 & -0.25 & -0.36 & -0.31
    \\
    {} & {} & 3 & 1.51 & 5.34 & 3.43 & -0.24 & -0.33 & -0.29
    \\ \cmidrule(lr){2-9}
    {} & \multirow{3}{*}{si} & 1 & 0.01 & 0.08 & 0.04 & -1.63 & -1.05 & -1.34
    \\
    {} & {} & 2 & 0.00 & 0.02 & 0.01 & -1.31 & -1.28 & -1.30
    \\
    {} & {} & 3 & 0.01 & 0.05 & 0.03 & -1.40 & -1.15 & -1.28
    \\
    \midrule[0.8\lightrulewidth]
    \multirow{12}{*}{T2I-EC} & \multirow{3}{*}{zh} & 1 & 19.25 & 11.91 & 15.58 & 0.06 & 0.18 & 0.12
    \\
    {} & {} & 2 & 0.09 & 0.11 & 0.10 & -1.75 & -1.60 & -1.68
    \\
    {} & {} & 3 & 18.60 & 12.27 & 15.43 & 0.05 & 0.18 & 0.11
    \\ \cmidrule(lr){2-9}
    {} & \multirow{3}{*}{de} & 1 & 17.91 & 25.72 & 21.81 & 0.18 & 0.38 & 0.28
    \\
    {} & {} & 2 & 18.64 & 26.20 & 22.42 & 0.19 & 0.41 & 0.30
    \\
    {} & {} & 3 & 18.56 & 25.82 & 22.19 & 0.19 & 0.39 & 0.29
    \\ \cmidrule(lr){2-9}
    {} & \multirow{3}{*}{ne} & 1 & 0.06 & 0.03 & 0.04 & -1.27 & -1.14 & -1.20
    \\
    {} & {} & 2 & 0.02 & 0.11 & 0.07 & -1.33 & -1.06 & -1.20
    \\
    {} & {} & 3 & 2.36 & 5.92 & 4.14 & -0.20 & -0.27 & -0.24
    \\ \cmidrule(lr){2-9}
    {} & \multirow{3}{*}{si} & 1 & 1.10 & 4.33 & 2.72 & -0.25 & -0.29 & -0.27
    \\
    {} & {} & 2 & 0.01 & 0.19 & 0.10 & -1.42 & -1.12 & -1.27
    \\
    {} & {} & 3 & 1.28 & 4.76 & 3.02 & -0.18 & -0.27 & -0.22
    \\
    \bottomrule
  \end{tabular}
  \caption{Full results of our main experiment with ResNet image representation.}
  \label{tab:main-results-resnet-full}
\end{table*}
\begin{table*}[ht!]
  \centering
  \begin{tabular}{lcccccccc}
    \toprule
    \multirow{2}{*}{Model} & \multirow{2}{*}{Language} & \multirow{2}{*}{Seed} & \multicolumn{3}{c}{BLEU} & \multicolumn{3}{c}{COMET}
    \\ \cmidrule(lr){4-6} \cmidrule(lr){7-9}
    {} & {} & {} & en$\to$X & X$\to$en & mean & en$\to$X & X$\to$en & mean
    \\
    \midrule
    \multirow{12}{*}{baseline (mBART + BT)} & \multirow{3}{*}{zh} & 1 & 17.21 & 11.35 & 14.28 & -0.04 & 0.14 & 0.05
    \\
    {} & {} & 2 & 18.38 & 11.39 & 14.89 & 0.02 & 0.14 & 0.08
    \\
    {} & {} & 3 & 18.45 & 11.36 & 14.90 & 0.03 & 0.15 & 0.09
    \\ \cmidrule(lr){2-9}
    {} & \multirow{3}{*}{de} & 1 & 18.66 & 25.83 & 22.24 & 0.18 & 0.39 & 0.29
    \\
    {} & {} & 2 & 19.06 & 25.73 & 22.39 & 0.20 & 0.38 & 0.29
    \\
    {} & {} & 3 & 18.79 & 25.88 & 22.33 & 0.22 & 0.40 & 0.31
    \\ \cmidrule(lr){2-9}
    {} & \multirow{3}{*}{ne} & 1 & 1.94 & 4.74 & 3.34 & -0.19 & -0.36 & -0.27
    \\
    {} & {} & 2 & 1.84 & 4.94 & 3.39 & -0.20 & -0.34 & -0.27
    \\
    {} & {} & 3 & 2.14 & 5.07 & 3.60 & -0.24 & -0.34 & -0.29
    \\ \cmidrule(lr){2-9}
    {} & \multirow{3}{*}{si} & 1 & 1.29 & 4.53 & 2.91 & -0.29 & -0.31 & -0.30
    \\
    {} & {} & 2 & 1.18 & 4.73 & 2.95 & -0.18 & -0.28 & -0.23
    \\
    {} & {} & 3 & 1.21 & 4.35 & 2.78 & -0.20 & -0.32 & -0.26
    \\
    \midrule[0.8\lightrulewidth]
    \multirow{12}{*}{I2I-EC} & \multirow{3}{*}{zh} & 1 & 16.66 & 10.94 & 13.80 & -0.07 & 0.13 & 0.03
    \\
    {} & {} & 2 & 17.46 & 10.87 & 14.16 & -0.01 & 0.13 & 0.06
    \\
    {} & {} & 3 & 18.84 & 11.64 & 15.24 & 0.03 & 0.16 & 0.10
    \\ \cmidrule(lr){2-9}
    {} & \multirow{3}{*}{de} & 1 & 18.64 & 26.17 & 22.40 & 0.22 & 0.40 & 0.31
    \\
    {} & {} & 2 & 17.98 & 25.20 & 21.59 & 0.20 & 0.38 & 0.29
    \\  
    {} & {} & 3 & 18.09 & 25.35 & 21.72 & 0.20 & 0.40 & 0.30
    \\ \cmidrule(lr){2-9}
    {} & \multirow{3}{*}{ne} & 1 & 1.02 & 4.68 & 2.85 & -0.41 & -0.38 & -0.39
    \\
    {} & {} & 2 & 1.87 & 5.19 & 3.53 & -0.26 & -0.33 & -0.29
    \\
    {} & {} & 3 & 1.79 & 5.29 & 3.54 & -0.20 & -0.34 & -0.27
    \\ \cmidrule(lr){2-9}
    {} & \multirow{3}{*}{si} & 1 & 0.30 & 1.64 & 0.97 & -1.14 & -0.59 & -0.87
    \\
    {} & {} & 2 & 0.16 & 0.55 & 0.36 & -0.88 & -0.88 & -0.88
    \\
    {} & {} & 3 & 0.76 & 4.88 & 2.82 & -0.37 & -0.29 & -0.33
    \\
    \midrule[0.8\lightrulewidth]
    \multirow{12}{*}{T2I-EC} & \multirow{3}{*}{zh} & 1 & 0.04 & 0.09 & 0.07 & -1.69 & -1.43 & -1.56
    \\
    {} & {} & 2 & 17.77 & 12.02 & 14.90 & 0.00 & 0.18 & 0.09
    \\
    {} & {} & 3 & 17.24 & 11.23 & 14.24 & -0.03 & 0.13 & 0.05
    \\ \cmidrule(lr){2-9}
    {} & \multirow{3}{*}{de} & 1 & 10.45 & 14.14 & 12.29 & -0.42 & -0.30 & -0.36
    \\
    {} & {} & 2 & 18.52 & 25.93 & 22.23 & 0.20 & 0.40 & 0.30
    \\
    {} & {} & 3 & 18.26 & 25.61 & 21.94 & 0.19 & 0.38 & 0.28
    \\ \cmidrule(lr){2-9}
    {} & \multirow{3}{*}{ne} & 1 & 0.75 & 2.49 & 1.62 & -0.85 & -0.58 & -0.71
    \\
    {} & {} & 2 & 0.09 & 0.07 & 0.08 & -1.37 & -1.18 & -1.28
    \\
    {} & {} & 3 & 0.02 & 0.04 & 0.03 & -1.35 & -1.17 & -1.26
    \\ \cmidrule(lr){2-9}
    {} & \multirow{3}{*}{si} & 1 & 0.02 & 0.15 & 0.09 & -2.00 & -1.45 & -1.72
    \\
    {} & {} & 2 & 0.04 & 0.19 & 0.12 & -2.02 & -1.32 & -1.67
    \\
    {} & {} & 3 & 1.05 & 4.18 & 2.61 & -0.33 & -0.28 & -0.30
    \\
    \bottomrule
  \end{tabular}
  \caption{Full results of our main experiment with CLIP-Large image representation.}
  \label{tab:main-results-clipl-full}
\end{table*}
\begin{table*}[ht!]
    \centering
    \begin{tabular}{lcccccccc}
    \toprule
    \multirow{2}{*}{Manipulation} & \multirow{2}{*}{Language} & \multirow{2}{*}{Seed} & \multicolumn{3}{c}{BLEU} & \multicolumn{3}{c}{COMET}
    \\ \cmidrule(lr){4-6} \cmidrule(lr){7-9}
    {} & {} & {} & en$\to$X & X$\to$en & mean & en$\to$X & X$\to$en & mean
    \\
    \midrule
    \multirow{6}{*}{CLIP-img} & \multirow{3}{*}{de} & 1 & 10.45 & 14.14 & 12.29 & -0.42 & -0.30 & -0.36
    \\
    {} & {} & 2 & 18.52 & 25.93 & 22.23 & 0.20 & 0.40 & 0.30
    \\
    {} & {} & 3 & 18.26 & 25.61 & 21.94 & 0.19 & 0.38 & 0.28
    \\
    \cmidrule(lr){2-9}
    {} & \multirow{3}{*}{si} & 1 & 0.02 & 0.15 & 0.09 & -2.00 & -1.45 & -1.72
    \\
    {} & {} & 2 & 0.04 & 0.19 & 0.12 & -2.02 & -1.32 & -1.67
    \\
    {} & {} & 3 & 1.05 & 4.18 & 2.61 & -0.33 & -0.28 & -0.30
    \\
    \midrule
    \multirow{6}{*}{Init BT} & \multirow{3}{*}{de} & 1 & 18.49 & 25.87 & 22.18 & 0.19 & 0.40 & 0.30
    \\
    {} & {} & 2 & 17.28 & 24.89 & 21.08 & 0.12 & 0.32 & 0.22
    \\
    {} & {} & 3 & 18.20 & 25.39 & 21.80 & 0.22 & 0.40 & 0.31
    \\
    \cmidrule(lr){2-9}
    {} & \multirow{3}{*}{si} & 1 & 0.94 & 4.56 & 2.75 & -0.43 & -0.27 & -0.35
    \\
    {} & {} & 2 & 1.24 & 4.84 & 3.04 & -0.28 & -0.25 & -0.27
    \\
    {} & {} & 3 & 0.09 & 0.62 & 0.35 & -1.24 & -0.84 & -1.04
    \\
    \midrule
    \multirow{6}{*}{Interleave} & \multirow{3}{*}{de} & 1 & 18.23 & 25.56 & 21.90 & 0.15 & 0.39 & 0.27
    \\
    {} & {} & 2 & 18.29 & 25.69 & 21.99 & 0.18 & 0.38 & 0.28
    \\
    {} & {} & 3 & 17.93 & 25.81 & 21.87 & 0.16 & 0.39 & 0.27
    \\
    \cmidrule(lr){2-9}
    {} & \multirow{3}{*}{si} & 1 & 0.01 & 0.02 & 0.02 & -1.57 & -1.34 & -1.46
    \\
    {} & {} & 2 & 1.25 & 4.84 & 3.05 & -0.34 & -0.25 & -0.30
    \\
    {} & {} & 3 & 1.04 & 4.37 & 2.70 & -0.46 & -0.28 & -0.37
    \\
    \bottomrule
    \end{tabular}
\caption{Results from several T2I-EC training pipeline manipulations.}
\label{tab:ablation-results-full}
\end{table*}
\end{document}